\pdfoutput = 1

\documentclass[10pt,twocolumn,letterpaper]{article}

\usepackage{cvpr}
\makeatletter
\@namedef{ver@everyshi.sty}{}
\makeatother
\usepackage{tikz}

\usepackage{times}
\usepackage{epsfig}
\usepackage{graphicx}
\usepackage{amsmath}
\usepackage{amssymb}
\DeclareMathOperator*{\argmax}{arg\!max}

\usepackage{xcolor}
\usepackage{caption}
\usepackage{subcaption}
\usepackage{array}
\usepackage{tabulary}
\usepackage{algpseudocode,algorithm,algorithmicx}

\usepackage[pagebackref=true,breaklinks=true,letterpaper=true,colorlinks,bookmarks=false]{hyperref}

\cvprfinalcopy 
\def\cvprPaperID{7} 

\ifcvprfinal\fi
\begin{document}

\title{Video Captioning via Hierarchical Reinforcement Learning}

\author{
Xin Wang, \quad Wenhu Chen, \quad Jiawei Wu, \quad Yuan-Fang Wang, \quad William Yang Wang\\
\\
University of California, Santa Barbara \\
{\tt\small \{xwang,wenhuchen,jiawei\_wu,yfwang,william\}@cs.ucsb.edu}
}

\maketitle

\begin{abstract}
Video captioning is the task of automatically generating a textual description of the actions in a video. Although previous work (\textit{e.g.} sequence-to-sequence model) has shown promising results in abstracting a coarse description of a short video, it is still very challenging to caption a video containing multiple fine-grained actions with a detailed description. This paper aims to address the challenge by proposing a novel hierarchical reinforcement learning framework for video captioning, where a high-level Manager module learns to design sub-goals and a low-level Worker module recognizes the primitive actions to fulfill the sub-goal. With this compositional framework to reinforce video captioning at different levels, our approach significantly outperforms all the baseline methods on a newly introduced large-scale dataset for fine-grained video captioning. Furthermore,  our non-ensemble model has already achieved the state-of-the-art results on the widely-used MSR-VTT dataset.

\end{abstract}

\section{Introduction}
\begin{figure}[t]
\begin{center}
\includegraphics[width=3.2in]{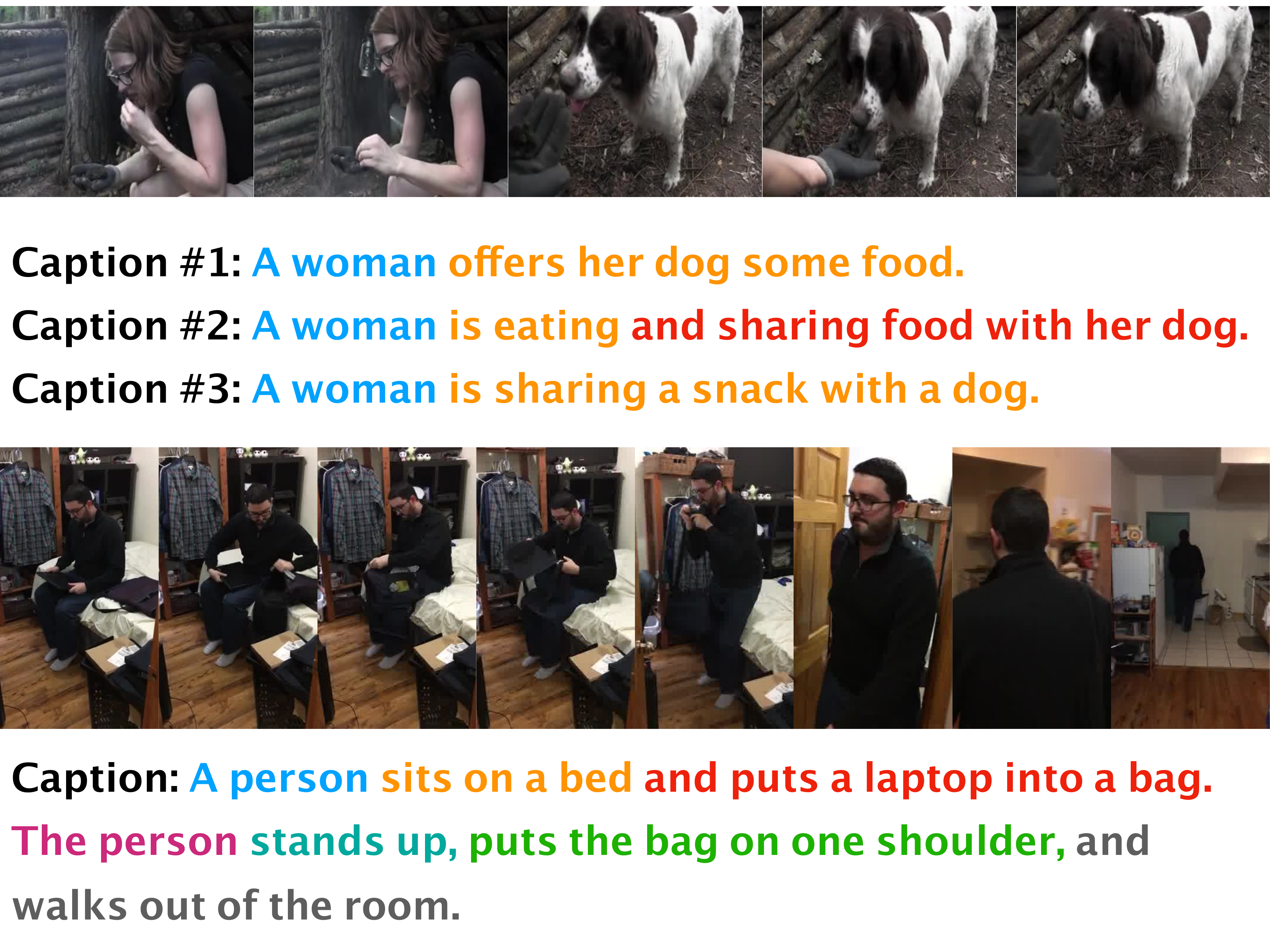}
\end{center}
\vspace{-2.5ex}
   \caption{Video captioning examples. Top row is an example from MSR-VTT dataset~\cite{xu2016msr}, which is summarized by three single captions. Bottom row is an example from Charades~\cite{sigurdsson2016hollywood} dataset, which consists of several dependent human activities and is described by multiple long sentences of complex structure.}
\vspace{-2ex}
\label{fig:intro}
\end{figure}

For most people, watching a brief video and describing what happened (in words) is an easy task. For machines, extracting the meaning from video pixels and generating natural-sounding description is a very challenging problem. However, due to its wide range of applications such as intelligent video surveillance and assistance to visually-impaired people, video captioning has drawn increasing attention from the computer vision community recently. Different from {\it image} captioning which aims at describing a static scene, {\it video} captioning is more challenging in the sense that a series of coherent scenes need to be understood in order to jointly generate multiple description segments (\textit{e.g.}, see Figure~\ref{fig:intro}).

Current video captioning tasks can mainly be divided into two families, single-sentence generation~\cite{xu2016msr,pan2016jointly} and paragraph generation~\cite{rohrbach14gcpr}. 
Single-sentence generation tends to abstract a whole video to a simple and high-level descriptive sentence, while paragraph generation tends to grasp more detailed actions, and generates multiple sentences of descriptions. 
However, even for paragraph generation, the paragraph is often split into multiple, single-sentence generation scenarios associated with ground truth temporal video intervals. 

In many practical cases, human activities are too complex to be described with short, simple sentences, and the temporal intervals are hard to be predicted ahead of time without a good understanding of the linguistic context.
For instance, in the bottom example of Figure~\ref{fig:intro},
there are five human actions in total: \textit{sit on a bed}, \textit{put a laptop into a bag} are happening simultaneously, and then followed by \textit{stand up}, \textit{put the bag on one shoulder} and \textit{walk out of the room} in order.
Such fine-grained caption requires a subtle and expressive mechanism to capture the temporal dynamics of the video content and associate that with semantic representations in natural language. 

In order to tackle this issue, we propose a ``divide and conquer" solution, which first divides a long caption into many small text segments (\textit{e.g.} different segments are in different colors as shown in Figure~\ref{fig:intro}), and then employs a sequence model to conquer each segment. Instead of forcing the sequence model to generate the whole sequence in one shot, we propose to guide the model to generate sentences segment by segment. With a higher-level sequence model designing the context of each segment, the low-level sequence model follows the guidance to generate the segment word by word. 

\begin{figure*}
\vspace*{-1ex}
\begin{center}
\includegraphics[width=5.5in]{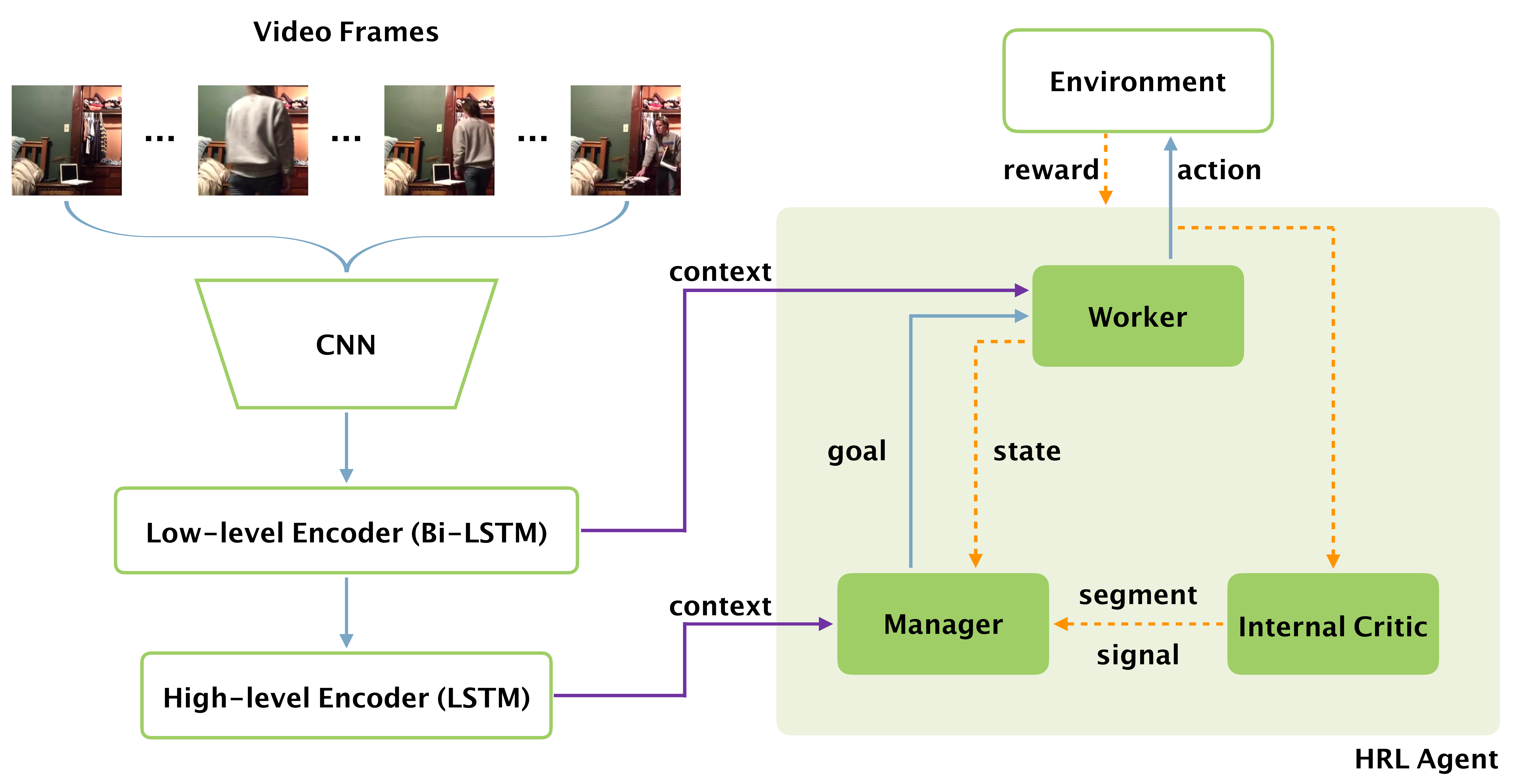}  
\end{center}
\vspace*{-2ex}
   \caption{Overview of the HRL framework for video captioning. Please see Sec.~\ref{sec:overview} for explanation.}
\label{fig:overview}
\vspace*{-1ex}
\end{figure*}

In this paper, we propose a novel hierarchical reinforcement learning (HRL) framework to realize this two-level mechanism. The textual and video context can be viewed as the reinforcement learning \textbf{environment}. Our framework is a fully-differentiable deep neural network (see Figure~\ref{fig:overview}) and consists of (1) the higher-level sequence model \textbf{manager} that sets \textbf{goals}  at a lower temporal resolution, (2) the lower-level sequence model \textbf{worker} that selects primitive \textbf{actions} at every time step by following the goals from the Manager, and (3) an \textbf{internal critic} that determines whether a goal is accomplished or not. More specifically, by exploiting the context from both the environment and finished goals, the manager emits a new goal for a new segment, and the worker receives the goal as guidance to generate the segment by producing words sequentially. Moreover, the internal critic is employed to evaluate whether the current textual segment is accomplished. 

Furthermore, we equip both the manager and worker with an attention module over the video features (Sec~\ref{sec:attention}) to introduce hierarchical attention internally so that the manager will focus on a wider range of temporal dynamics while the worker's attention is narrowed down to local dynamics conditioned on the goals. 
To the best of our knowledge, this is the first work that strives to develop a hierarchical reinforcement learning approach to reinforce video captioning at different levels. Our main contributions are four-fold:
\begin{itemize}
\item We propose a hierarchical deep reinforcement learning framework to efficiently learn the semantic dynamics when captioning a video.
\item We formulate an alternative, novel training approach over stochastic and deterministic policy gradient.
\item We introduce a new large-scale dataset for fine-grained video captioning, Charades Captions\footnote{Charades Captions was obtained by preprocessing the raw Charades dataset~\cite{sigurdsson2016hollywood}. The processed Charades Captions dataset can be downloaded here: \url{http://www.cs.ucsb.edu/~xwang/data/CharadesCaptions.zip}}, and validate the effectiveness of the proposed method in it.
\item We further evaluate our approach on MSR-VTT dataset and achieve the state-of-the-art results even when training on a single type of features. 
\end{itemize}

\section{Related Work}
\paragraph{Video Captioning}
S2VT~\cite{venugopalan2015sequence} first generalized LSTM to video captioning and proposed a sequence-to-sequence model for it. Since then, numerous improvements were introduced, such as attention~\cite{yao2015describing,yu2016video_att}, hierarchical recurrent neural network (RNN)~\cite{yu2016video,pan2016hierarchical,baraldi2016hierarchical,song2017hierarchical,wang2018watch}, C3D features~\cite{Shen_2017_CVPR}, joint embedding space~\cite{ramanishka2016multimodal}, language fusion~\cite{gan2016semantic}, multi-task learning~\cite{pasunuru2017multi}, etc. But most of them use the maximum-likelihood algorithm, which maximizes the probability of current ground-truth output given previous ground-truth output, while the previous ground-truth is in general unknown during test time. This inconsistency issue known as exposure bias has largely hindered the system performance.

In order to address the inconsistency issue, Ranzato \textit{et al.}~\cite{ranzato2015sequence} proposed to directly optimize non-differentiable metric scores using the REINFORCE algorithm~\cite{williams1992simple}. But the problem persisted that the expected gradient computed using policy gradient typically exhibited high variance and was often unstable without proper context-dependent normalization. Naturally, the variance could be reduced by adding a baseline~\cite{liu2016improved,rennie2016self} or even an actor-critic method that trained an additional critic to estimate the value of each generated word~\cite{bahdanau2016actor,Ren_etal_CVPR_17,zhang2017actor}. Pasunuru and Bansal~\cite{pasunuru2017reinforced} applied policy gradient with baseline on video captioning and presented textual entailment loss to adjust the CIDEr reward. Unfortunately, these previous work for image/video captioning fail to grasp the high-level semantic flow. Our HRL model aims to address this issue with a hierarchical reinforcement learning framework.

Another line of work is dense video captioning~\cite{krishna2017dense}, which focuses on detecting multiple events that occur in a video and describing each of them. But it does not aim to solve the single-sentence generation scenario. While our method aims to generate one or multiple sentences for a sequence of continuous actions (one or multiple). 

\paragraph{Hierarchical Reinforcement Learning} Recent work has revealed the effectiveness of hierarchical reinforcement learning frameworks on Atari games~\cite{kulkarni2016hierarchical,vezhnevets2017feudal}. Peng \textit{et al.} built a composite dialogue policy using hierarchical Q-learning to fulfill complex dialogue tasks like traveling plans \cite{peng2017composite}. In the typical HRL setting, there was a high-level agent that operated at the lower temporal resolution to set a sub-goal, and a low-level agent that selected primitive actions by following the sub-goal from the high-level agent. Our proposed HRL framework for video captioning is aligned to these studies but has a key difference from the typical HRL setting: instead of having the internal critic to provide an intrinsic reward to encourage the low-level agent to accomplish the sub-goal, we focus on exploiting the extrinsic rewards in different time spans. Besides, we are the first to consider HRL in the intersection vision and language.

\section{Our Approach}
\label{sec:method}

\subsection{Overview} 
\label{sec:overview}
Our proposed HRL framework follows the general encoder-decoder framework (see Figure~\ref{fig:overview}). In the encoding stage, video frame features $v=\{v_i\}$ are first extracted by a pretrained convolutional neural network (CNN)~\cite{krizhevsky2012imagenet} model, where $i \in \{1, ..., n\}$ indexes the frames in the temporal order. Then the frame features are passed through a low-level Bi-LSTM\footnote{Bidirectional long short-term memory~\cite{schuster1997bidirectional}} encoder and a high-level LSTM\footnote{Long short-term memory~\cite{hochreiter1997long}} encoder successively to obtain low-level encoder output $h^{E_w} = \{h^{E_w}_i\}$ ($E_w$ denotes the encoder associated with the Worker), and high-level encoder output $h^{E_m} = \{h^{E_m}_i\}$ ($E_m$ denoting the encoder associated with the Manager), where $i \in \{1, ..., n\}$. In the decoding stage, our HRL agent plays the role of a decoder, and outputs a language description $a_1 a_2 ... a_T \in V^T$, where $T$ is the length of the generated caption and $V$ is the vocabulary set.

The HRL agent is composed of three components: a low-level \textit{worker}, a high-level \textit{manager}, and an \textit{internal critic}. The manager operates at a lower temporal resolution and emits a goal when needed for the worker to accomplish, and the worker generates a word for each time step by following the goal proposed by the manager. 
In other words, the manager asks the worker to generate a semantic segment, and the worker generates the corresponding words in the next few time steps in order to fulfill the job. 
The internal critic determines if the worker has accomplished the goal and sends a binary segment signal to the manager to help it update goals. The whole pipeline terminates once an end of sentence token is reached.

\subsection{Policy Network} \label{sec:policy_network}

\paragraph{Attention Module} \label{sec:attention}
As mentioned above, the CNN-RNN encoder receives the video inputs to generate a sequence of vectors $h^{E_w} = \{h^{E_w}_i\}$ and $h^{E_m} = \{h^{E_m}_i\}$. One may directly take them as the inputs to the worker and the manager. We instead adopt an attention mechanism to better capture the temporal dynamics, and form the context vector for their use. In our model, both the manager and the worker are equipped with an attention module.

\begin{figure}[t]
\begin{center}
\includegraphics[width=3.5in]{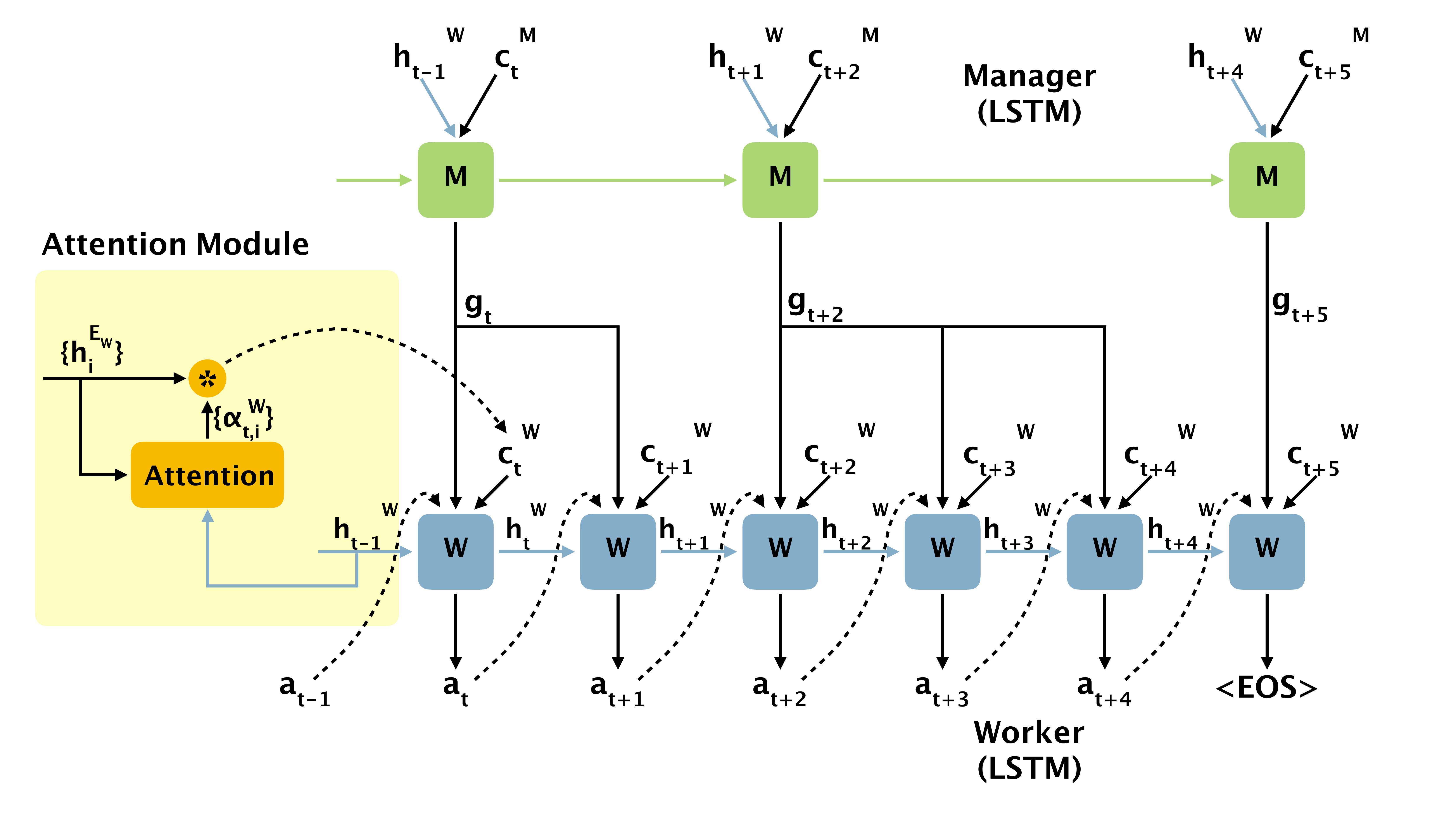}  
\end{center}
\vspace*{-3ex}
   \caption{An example of the unrolled HRL agent in the decoding stage (from time step $t$ to $t+5$). The yellow region shows how the attention module is incorporated into the encoder-decoder framework.}
\label{fig:hrl_agent}
\vspace*{-1ex}
\end{figure}

The left-hand side of Figure~\ref{fig:hrl_agent} is a demo attention module for the worker, at each time step $t$, the context vector $c_t^W$ is computed as a weighted sum over the encoder's all hidden states $\{h_i^{E_w}\}$
\begin{equation} \label{eq:context}
    c_t^W = \sum \alpha^W_{t,i} h_i^{E_w}
\end{equation}
These attention weights $\{\alpha^W_{t,i}\}$ act as an alignment mechanism by giving higher weights to certain encoder hidden states which match the worker's current status, and are defined as
\begin{equation} \label{eq:att1}
    \alpha^W_{t,i} = \frac{\exp(e_{t,i})}{\sum_{k=1}^n \exp(e_{t,k})}
\end{equation}
where
\begin{equation} \label{eq:att2}
    e_{t,i} = w^T \tanh (W_a h_i^{E_w} + U_a h_{t-1}^W + b_a)
\end{equation}
where $w, W_a, U_a$ and $b_a$ are learned parameters; $h_{t-1}^W$ is the worker LSTM's hidden state at previous step. 

The manager's attention module follows the same paradigm as the worker's, which can be described by replacing the corresponding terms in Equation~\ref{eq:context}, \ref{eq:att1}, and \ref{eq:att2}.

\paragraph{Manager and Worker}

As is shown in Figure~\ref{fig:hrl_agent}, the concatenation of [$c_t^M, h_{t-1}^W$] is fed as the input to the manager LSTM to produce the semantically meaningful goal. With the help of the context and the sentence state at previous time steps, the manager can obtain the knowledge of the environment status. The output of the manager LSTM $h_t^M$ is then projected as a latent continuous goal vector $g_t$. Formally, 
\begin{equation}
    h_t^M = S^M(h_{t-1}^M, [c_t^M, h_{t-1}^W])
\end{equation}
\begin{equation}
    g_t = u_M(h_t^M)
\end{equation}
where $S^M$ denotes the non-linear function of the manager LSTM and $u_M$ is a function to project hidden states into goal space.

The worker receives the goal $g_t$, takes the concatenation of [$c_t^W, g_t, a_{t-1}$] as the input, and outputs the probabilities $\pi_t$ over all actions $a_t \in V$ after a series of computations:
\begin{equation}\label{eq:worker_lstm}
    h_t^W = S^W(h_{t-1}^W, [c_t^W, g_t, a_{t-1}])
\end{equation}
\begin{equation}\label{eq:linear_w}
    x_t = u_W(h_t^W)
\end{equation}
\begin{equation}\label{eq:softmax}
    \pi_t = SoftMax(x_t) 
\end{equation}
where $S^W$ is the non-linear function of the worker LSTM and $u_W$ is a also a function to project hidden states into the input to softmax layer.

\paragraph{Internal Critic}
In order to determine whether the worker has accomplished a goal $g_t$, we employ an internal critic to evaluate worker's progress. The internal critic uses an RNN structure, which takes a word sequence as the input to discriminate whether an end has been reached. Let $z_t$ denote the signal of internal critic and $h^I_t$ denote the hidden state of the RNN at time step $t$, formally we describe the probability $p(z_t)$ as follows:
\begin{align}
\begin{split}
h^I_t &= RNN(h^I_{t-1}, a_{t}) \\
p(z_t) &= sigmoid(W_z h^I_t + b_z) 
\end{split}
\end{align}
where $a_t$ is the action taken by the worker and $W_z, b_z$ denotes the parameters of the feed-forward neural network. In order to train the parameters of the linear layer and recurrent network, we propose to maximize the likelihood of given ground truth signal $\{z_t^*\}$:
\begin{align}
\argmax \sum_t \log p(z_t^*|a_1, \cdots, a_{t-1})
\end{align}
Once the critic model is optimized, we will fix it to service the usage of the manager. 

\subsection{Learning}
\label{sec:hier_policy_training}
As described in Sec.~\ref{sec:policy_network}, the manager policy is actually deterministic, which can be further denoted as $g_t = \mu_{\theta_m}(s_t)$ with $\theta_m$ representing the parameters of the manager, while the worker policy is a stochastic policy denoted by $a_t \sim \pi_{\theta_w}(a_t; s_t, g_t)$, where $\theta_w$ represents the parameters of the worker. The reason why the worker policy is stochastic is that its action $a_t$ is selecting a word from the vocabulary $V$. But for the manager, the generated goal is latent, which cannot be directly supervised. Thus with a deterministic manager policy, we can warm start both the manager and worker simultaneously by viewing them as a composite agent.   

In this section, we first derive the mathematical reinforce learning methods for the policies separately (Sec.~\ref{sec:worker_learning} and \ref{sec:manager_learning}), and then introduce the training algorithm of the proposed HRL method (Sec.~\ref{sec:training_algorithm}). We also discuss the reward definitions (Sec.~\ref{sec:reward}) and imitation learning of our HRL policy (Sec.~\ref{sec:imitation}).

\subsubsection{Stochastic Worker Policy Learning}
\label{sec:worker_learning}
We consider a standard reinforcement learning setup. At each step $t$, the worker select an action $a_t$ ($a_t \in V$) conditioned on $g_t$ from the manager. The environment responds with a new state $s_{t+1}$ and a scalar reward $r_t$. The process continues until a $<$EOS$>$ token is generated. The objective of the worker is to maximize the discounted return $R_t = \sum_{k=0}^\infty \gamma^k r_{t+k}$. Thus its loss function can be written as 
\begin{equation}
    L(\theta_w) = - \mathbb{E}_{a_t \sim \pi_{\theta_w}} [R(a_t)] 
\end{equation}
to minimize the negative expected reward function.
Based on REINFORCE algorithm~\cite{williams1992simple}, the gradient of non-differentiable, reward-based loss function can be derived as
\begin{equation}
    \nabla_{\theta_w} L(\theta_w) = - \mathbb{E}_{a_t \sim \pi_{\theta_w}} [R(a_t) \nabla_{\theta_w} \log \pi_{\theta_w} (a_t) ] 
\end{equation}
In practice $L(\theta_w)$ is typically estimated with a single sample from $\pi_{\theta_w}$:
\begin{equation}
    \nabla_{\theta_w} L(\theta_w) \approx - R(a_t) \nabla_{\theta_w} \log \pi_{\theta_w} (a_t) 
\end{equation}
The policy gradient given by REINFORCE can be further generalized to reduce the variance without changing the expected gradient, by subtracting the reward with a baseline~\cite{sutton1998reinforcement}:
\begin{equation} \label{PG_worker}
    \nabla_{\theta_w} L(\theta_w) \approx - (R(a_t) - b_t^w) \nabla_{\theta_w} \log \pi_{\theta_w} (a_t) 
\end{equation}
where $b_t^w$ is the estimated baseline, which can be a function of $\theta_w$ or $t$~\cite{ranzato2015sequence}. In our case, the baseline is estimated by a linear regressor with the worker's hidden state $h_t^W$ as the input. During back propagation, the gradient passing is cut off between the worker LSTM and the baseline estimator.

For a better understanding of the policy gradient, we can further derive the loss function using the chain rule
\begin{equation}
    \nabla_{\theta_w} L(\theta_w) = \sum_{t=1}^T \frac{\partial L}{\partial x_t} \frac{\partial x_t}{\partial \theta_w}
\end{equation}
where $x_t$ is the input to the \textit{SoftMax} layer (see Equation~\ref{eq:linear_w}). Using REINFORCE with baseline the estimation of $\frac{\partial L}{\partial x_t}$ is given by~\cite{zaremba2015reinforcement}:
\begin{equation}
    \frac{\partial L}{\partial x_t} = (R(a_t) - b_t^w)(\pi_{\theta_w}(a_t) - 1_{a_t})
\end{equation}
which means if the reward $R(a_t)$ of the sample word $a_t$ is greater than the baseline $b_t$, the gradient is negative and thus the model encourages the distribution by increasing the probability of the word, otherwise, it discourages the distribution accordingly.

\subsubsection{Deterministic Manager Policy Learning}
\label{sec:manager_learning}
The key to our HRL framework is to effectively learn the goal $g_t$  generated by the manager and then guides the worker to achieve the latent objective. But the difficulty of training the manager is that it does not directly interact with the environment because the action it takes is to produce a latent vector $g_t$ in a continuous high-dimensional space, which indirectly influences the environment by directing the Worker's behavior. Therefore, we are especially interested in coming up solutions to encourage the manager towards more effective caption generation.

Inspired by the deterministic policy gradient algorithms~\cite{silver2014deterministic, lillicrap2015continuous}, we propose to learn the deterministic policy $\mu_{\theta_m}(s_t)$ from trajectories generated by the stochastic worker policy $\pi_{\theta_w}(a_t; s_t, g_t)$. When training the target manager policy, we fix the worker policy as an Oracle behavior policy. More specifically, the manager outputs a goal $g_t$ at step $t$ and the worker then runs $c$ steps to generate the expected segment $e_{t,c} = a_t a_{t+1}...a_{t+c-1}$ by following the goal ($c$ is length of the generated segment). Since the worker is fixed as an Oracle behavior policy, we only need to consider the training of the manager. Then the environment responds with an new state $s_{t+c}$ and a scalar reward $r(e_{t,c})$. Thus the objective becomes minimizing the negative discounted return $R_t(e_{t,c})$, in formula 
\begin{equation}
    L(\theta_m) = - \mathbb{E}_{g_t}[R(e_t) \pi(e_{t,c}; s_t, g_t=\mu_{\theta_m}(s_t)]
\end{equation}
After applying the chain rule to the loss function with respect to the manager's parameters $\theta_m$, the manager is updated with
\begin{equation}
    \nabla_{\theta_m} L(\theta_m) = - \mathbb{E}_{g_t} [R(e_{t,c}) \nabla_{g_t} \pi(e_{t,c}; s_t, g_t) \nabla_{\theta_m} \mu_{\theta_m}(s_t) ]
\end{equation}
The above gradients can be approximated from a single sampled segment $e_{t,c}$ and after adopting policy gradient on the worker policy, 
\begin{equation} \label{eq:deter_raw}
    \nabla_{\theta_m} L(\theta_m) = - R(e_{t,c}) \nabla_{g_t} \log \pi(e_{t,c}) \nabla_{\theta_m} \mu_{\theta_m}(s_t)
\end{equation}
Since the worker LSTM is indeed a Markov decision process and the probability of the current action $a_t$ is conditioned on the action $a_{t-1}$ at previous step (see Equation~\ref{eq:worker_lstm},\ref{eq:linear_w},\ref{eq:softmax}), we have
\begin{equation} \label{eq:mdp}
\log \pi(e_{t,c}) = \log \pi(a_t..a_{t+c-1}) = \sum_{i=t}^{t+c-1} \log \pi(a_i)
\end{equation}
Combining Equation~\ref{eq:deter_raw} and \ref{eq:mdp}, then the gradients become
\begin{equation}
\nabla_{\theta_m} L(\theta_m) = - R(e_{t,c}) [\sum_{i=t}^{t+c-1} \nabla_{g_t} \log \pi(a_i)] \nabla_{\theta_m} \mu_{\theta_m}(s_t)
\end{equation}
The final gradients for the manager training is obtained by adding the baseline estimator to reduce the variance as follows: 
\begin{align} \label{DPG_manager}
\begin{split}
& \nabla_{\theta_m} L(\theta_m) = \\
& - (R(e_{t,c}) - b_t^m) [\sum_{i=t}^{t+c-1} \nabla_{g_t} \log \pi(a_i)] \nabla_{\theta_m} \mu_{\theta_m}(s_t)
\end{split}
\end{align}
where $b_t^m$ is the baseline estimator, which is a linear regressor with the manager's hidden state $h_t^M$ as the input. 

A major challenge of learning in continuous action spaces is exploration. We follow the known DDPG~\cite{lillicrap2015continuous} to construct an exploration policy $\mu '$ by adding perturbation $\epsilon$ sampled from a Gaussian distribution $\mathcal{N}$ to our manager policy
\begin{align}
\mu'(s_t) =& \mu_{\theta_m}(s_t) + \epsilon
\end{align}
and the variance of Gaussian noise can be chosen to suit the environment.

\subsubsection{Reward Definition}\label{sec:reward}
Recent work on image captioning~\cite{rennie2016self} has shown that CIDEr as a reward performs the best among the traditional evaluation metrics (\textit{e.g.} CIDEr, BLEU or METEOR) for image/video captioning and can gain improvement on all other metrics. In our model, we also use CIDEr score to compute the reward. But instead of directly using the final CIDEr score of the whole generated caption as the reward for each word $a_t$, we adopt delta CIDEr score as the immediate reward. Let $f(x)=\text{CIDEr}(sent + x) - \text{CIDEr}(sent)$, where $sent$ is the previous generated caption. Then the discounted return for the worker is
\begin{equation}
R(a_t) = \sum_{k=0}^\infty \gamma^k f(a_{t+k})
\end{equation}
where $k$ denotes the time step of the worker's temporal resolution, and the discounted return for the manager is
\begin{equation}
R(e_t) = \sum_{n=0}^\infty \gamma^n f(e_{t+n})
\end{equation}
where $n$ is the time step of the manager's lower temporal resolution. Note that our approach is not limited to CIDEr score, other reasonable rewards (\textit{e.g.} deltaBLEU~\cite{galley2015deltableu}) can also be applied to the HRL framework.

\subsubsection{Training Algorithm}
\label{sec:training_algorithm}
Above we illustrate the learning methods to train the manager and the worker. In Algorithm~\ref{alg:hrl_training} we present the pseudo-code of our HRL training algorithm for video captioning. The manager policy and the worker policy are trained alternately. Basically, when training the worker, we assume the manager is well-posed, so we disable the goal exploration and only update the worker policy according to Equation~\ref{PG_worker}; when training the manager, we treat the worker as the Oracle behavior policy, so we generate the caption by greedy decoding and only update the manager policy following Equation~\ref{DPG_manager}.   

During testing, goal exploration is disabled, and beam search is employed to generate the results. Only one forward pass is needed at test time.

\begin{algorithm} [t]
    \caption{HRL training algorithm}
    \label{alg:hrl_training}
    \begin{algorithmic}[1]
        \Require{Training pairs $<$video, GT caption$>$}
        \State Randomly initialize the model parameters $\theta$
        \State Load the pretrained CNN model and internal critic
        \For{iteration=1,M}
            \State Randomly sample a minibatch
            \If{Train-Worker}
            \State Disable the goal exploration 
            \State Run a forward pass to get the sampled caption $a_1 a_2 ... a_T$ 
            \State Calculate $R(a_t)$ for each $a_t$
            \State Freeze the manager
            \State Update the worker policy using Equation~\ref{PG_worker}
            \ElsIf{Train-Manager}
            \State Initialize a random process $\mathcal{N}$ for goal exploration 
            \State Run a forward pass to get the greedily decoded caption $e_1 e_2 ... e_n$
            \State Calculate $R(e_t)$ for each $e_t$
            \State Freeze the worker
            \State Update the manager policy using Equation~\ref{DPG_manager}
            \EndIf
        \EndFor
            
    \end{algorithmic}
\end{algorithm}

\subsubsection{Imitation Learning}
\label{sec:imitation}
A major challenge for a reinforcement learning agent to have good convergence property is that the agent must start with a good policy at the beginning stage. For our model, we apply the cross-entropy loss optimization to warm start both the worker and the manager simultaneously, where the manager is completely treated as the latent parameters.  $\theta$ be the parameters of the whole model and ${a_1^*, a_2^*, ..., a_T^*}$ be the ground-truth word sequence, then the cross-entropy loss is defined as
\begin{equation}
L(\theta) = - \sum_{t=1}^T \log(\pi_{\theta}(a_t^*; a_1^*,...,a_{t-1}^*))
\end{equation}

\section{Experimental Results}
\label{sec:experiments}

\subsection{Datasets}
\paragraph{MSR-VTT} 
MSR-VTT~\cite{xu2016msr} is a dataset for general video captioning, which is derived from a wide variety of video categories (7,180 videos from 20 general categories), and contains 10,000 video clips (6,513 for training, 497 for validation, and the remaining 2,990 for testing). Each video contains 20 human annotated reference captions collected by Amazon Mechanical Turk (AMT). 

\paragraph{Charades Captions}
Charades~\cite{sigurdsson2016hollywood} is a large-scale dataset composed of 9,848 videos of daily indoors activities collected through AMT. 267 different users were presented with a sentence script (\textit{e.g.} a person fixes the bed then throws pillow on it) that included objects and actions from a fixed vocabulary, and the users recorded a video following the script using provided objects and actions. The original dataset contains 66,500 temporal annotations for 157 action classes, 41,104 labels for 46 object classes, and 27,847 textual descriptions of the videos.

While the Charades dataset is mainly used for action recognition and segmentation, one should note that the collected textual descriptions are very detailed and depict the fine-grained human activities happening in long videos. 
Thus, we preprocessed the raw Charades dataset by combining the textual descriptions and sentence scripts verified through AMT\footnote{For example, the sentence script of a video can be \textit{A person is taking a picture of a light while sitting in a chair.}, and the textual description is \textit{A person in a bedroom appears to use their phone to film or take a picture of the light fixture on the ceiling.} The latter is usually more detailed.}, 
and built a new large-scale dataset for detailed video captioning -- Charades Captions, which consists of 6,963 videos for training, 500 for validation and 1,760 for testing. Each video clip is annotated by multiple (typically 2-5) captions. The captions are more detailed and longer than those of MSR-VTT (average caption length: 24.13 \textit{vs} 9.28 words), which is more suitable for fine-grained video captioning.

\paragraph{Caption Segmentation}
In order to train the internal critic that determines if a goal is accomplished, we preprocessed the ground truth captions of the training sets of both datasets by breaking each caption into multiple semantic chunks. We segmented the captions mainly based on the Noun Phrase (NP) and Verb Phrase (VP) labels provided by the constituency parsing results (We utilized the open source toolkits Stanford CoreNLP\footnote{\url{https://stanfordnlp.github.io/CoreNLP/}}~\cite{manning-EtAl:2014:P14-5} and NLTK\footnote{\url{http://www.nltk.org}} for constituency parsing). 
For instance, the caption \textit{The person then tidies his area after he is done eating} was segmented into three sub-phrases, \textit{The person}, \textit{then tidies his area} and \textit{after he is done eating} with labels NP, VP and VP respectively. 
However, all we need to train the internal critic were the chunks, and labels were not used. 

\begin{table} 
\small
\begin{center}
  \begin{tabular}{ l | c c c c }
  
   Method & BLEU@4 & METEOR & ROUGE-L & CIDEr \\
   \hline\hline
   Mean-Pooling     & 30.4 & 23.7 & 52.0 & 35.0 \\
   Soft-Attention     &  28.5 & 25.0 & 53.3 & 37.1 \\
   S2VT                & 31.4 & 25.7 & 55.9 & 35.2 \\
   \hline
   v2t\_navigator     & 40.8 & 28.2 & 60.9 & 44.8  \\
   Aalto             & 39.8 & 26.9 & 59.8 & 45.7  \\
   VideoLAB         & 39.1 & 27.7 & 60.6 & 44.1    \\
   \hline
   XE-baseline         & \textbf{41.3} & 27.6 & 59.9 & 44.7 \\
   RL-baseline         & 40.6 & 28.5 & 60.7 & 46.3 \\ 
   HRL (Ours)        & \textbf{41.3} & \textbf{28.7} & \textbf{61.7} & \textbf{48.0} \\
   
  \end{tabular}
\end{center}
\vspace{-2ex}
\caption{Comparison with state of the arts on MSR-VTT dataset.}
\label{table:msrvtt}
\vspace{-1ex}
\end{table}

\subsection{Experimental Setup}
\paragraph{Evaluation Metrics} 
We adopted four diverse automatic evaluation metrics: BLEU, METEOR, ROUGE-L, and CIDEr-D. We used the standard evaluation code from MS-COCO server~\cite{chen2015microsoft} to obtain the results.  

\paragraph{Training Details} All the hyperparameters were tuned on the validation set. For both datasets, we sampled each video at $3 fps$ and extracted ResNet-152 features~\cite{he2016deep} from these sampled frames without fine-tuning. More training details can be found in the supplementary material.

\subsection{Results and Analysis}
\paragraph{Comparison with state of the arts on MSR-VTT}
In Table~\ref{table:msrvtt}, we compared our single-sentence captioning results with the-state-of-the-art methods on MSR-VTT dataset. We listed the results of Mean-Pooling~\cite{venugopalan:naacl15}, Soft-Attention~\cite{yao2015describing} and S2VT~\cite{venugopalan2015sequence} as reported in previous work~\cite{Shen_2017_CVPR}. We also compared with the top-3 results from MSR-VTT challenge, including v2t navigator~\cite{jin2016describing}, Aalto~\cite{shetty2016frame}, VideoLAB~\cite{ramanishka2016multimodal}. 

We implemented two baseline methods: an attention-based sequence-to-sequence model trained with cross-entropy loss (\textit{XE-baseline}), and the same model trained with policy gradient and CIDEr score as the RL reward (\textit{RL-baseline}). As shown in Table~\ref{table:msrvtt}, our XE-baseline achieved comparable results with the state-of-the-art results, and our RL-baseline further improved on all metrics. Moreover, our novel HRL method outperformed all the other algorithms listed in the table, which proved the effectiveness of our proposed method. 

\paragraph{Result Analysis on Charades Captions}
\begin{table} 
\small
\begin{center}
  \begin{tabular}{ l | c @{\hspace{0.2cm}} c @{\hspace{0.2cm}}c @{\hspace{0.2cm}}c @{\hspace{0.2cm}}c @{\hspace{0.2cm}}c @{\hspace{0.2cm}}c }
  
   Method & B@1 &B@2 &B@3 & B@4 & M & R & C \\
   \hline\hline
   XE-baseline     & 55.0  & 36.4 & 23.6 &  15.0 & 18.7&  39.0 & 16.7   \\
   RL-baseline     & 57.6 & 41.4 & 28.0 & \textbf{18.8} &  17.7 & 39.8 & 21.6 \\
   \hline
   HRL-16 & \textbf{64.4} & \textbf{44.3} & \textbf{29.4} & \textbf{18.8} & \textbf{19.5} & \textbf{41.4} & 23.2 \\
   HRL-32 & 64.0 &  43.4& 28.4 & 17.9 & 19.2 & 41.0 & 21.3  \\
   HRL-64 & 61.7 & 43.0 & 28.8 & \textbf{18.8} & 18.7 & 31.2 & \textbf{23.6} \\
   
  \end{tabular}
\end{center}
\vspace{-2ex}
\caption{Results on Charades Captions dataset. We reported BLEU (B), METEOR (M), ROUGH-L (R) and CIDEr (C) scores of our HRL method and two baselines for comparison.}
\label{table:charades}
\vspace{-1ex}
\end{table}

\begin{figure*}[t]
\begin{center}
\includegraphics[width=6.8in]{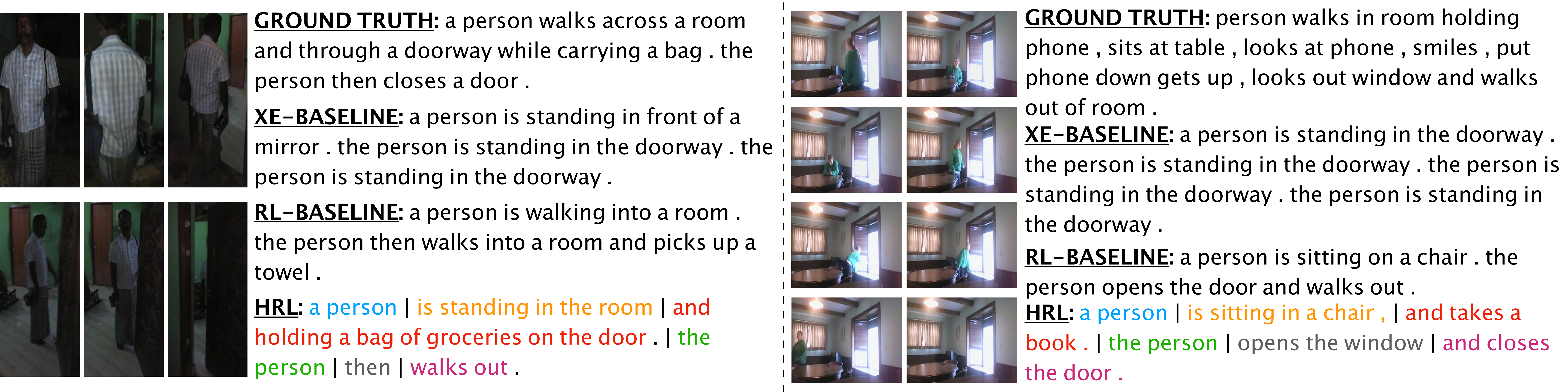}  
\end{center}
\vspace*{-2ex}
   \caption{Qualitative comparison with the baseline methods. The given examples were from the test set of the Charades Caption dataset.}
\label{fig:case_study}
\end{figure*}

Since there were no other papers reporting results on Charades Captions, we mainly compared our HRL model with our implementation of XE-baseline and RL-baseline. Meanwhile, we explored the dimension of the latent goal vector (We used HRL-$X$ to denote the HRL model with a goal dimension of $X$). As can be observed from Table~\ref{table:charades}, all our HRL models outperformed the baseline methods and brought significant improvements in different evaluation metrics. Note that our HRL model achieved bigger improvement over the baseline methods on Charades Captions dataset than on MSR-VTT. Given that fact that the average cation length of Charades Captions was much longer than that of MST-VTT (24.13 \textit{vs} 9.28 words), the difference of the improvement gaps demonstrated that our HRL model can gain better improvement on detailed descriptions of longer videos.

Among the HRL models, HRL-16 achieved the best on almost all metrics (CIDEr score was the second-best and slightly worse than HRL-64). Even though HRL-64 obtained better results on BLEU@4 and CIDEr, its results on other metrics were worse than HRL-32 (the ROUGE-L score was much lower than HRL-32). Thus, comparing the results of different HRL models, we could conclude that HRL-16 $>$ HRL-32 $\geq$ HRL-64. This result accorded with our speculation: higher dimension does not guarantee better performance, conversely, the exploration space grows exponentially as the dimension increases, making the learning even harder. A latent vector of small dimension like 16 is able to represent the semantically meaningful goal well.  

\paragraph{Qualitative Comparison with Baseline Methods} 
In Figure~\ref{fig:case_study}, we illustrated two examples from Charades Captions test set. According to the captions generated by different models, it is obvious that the generated results of our HRL model matched the ground truth captions better than the baseline methods. Moreover, due to the segment-by-segment generation manner, our HRL model was able to output a sequence of semantically meaningful phases (different phases were in different colors and segmented by ``$|$" as in Figure~\ref{fig:case_study}). 

\begin{figure}
\vspace*{-1ex}
\begin{center}
\includegraphics[width=3.3in]{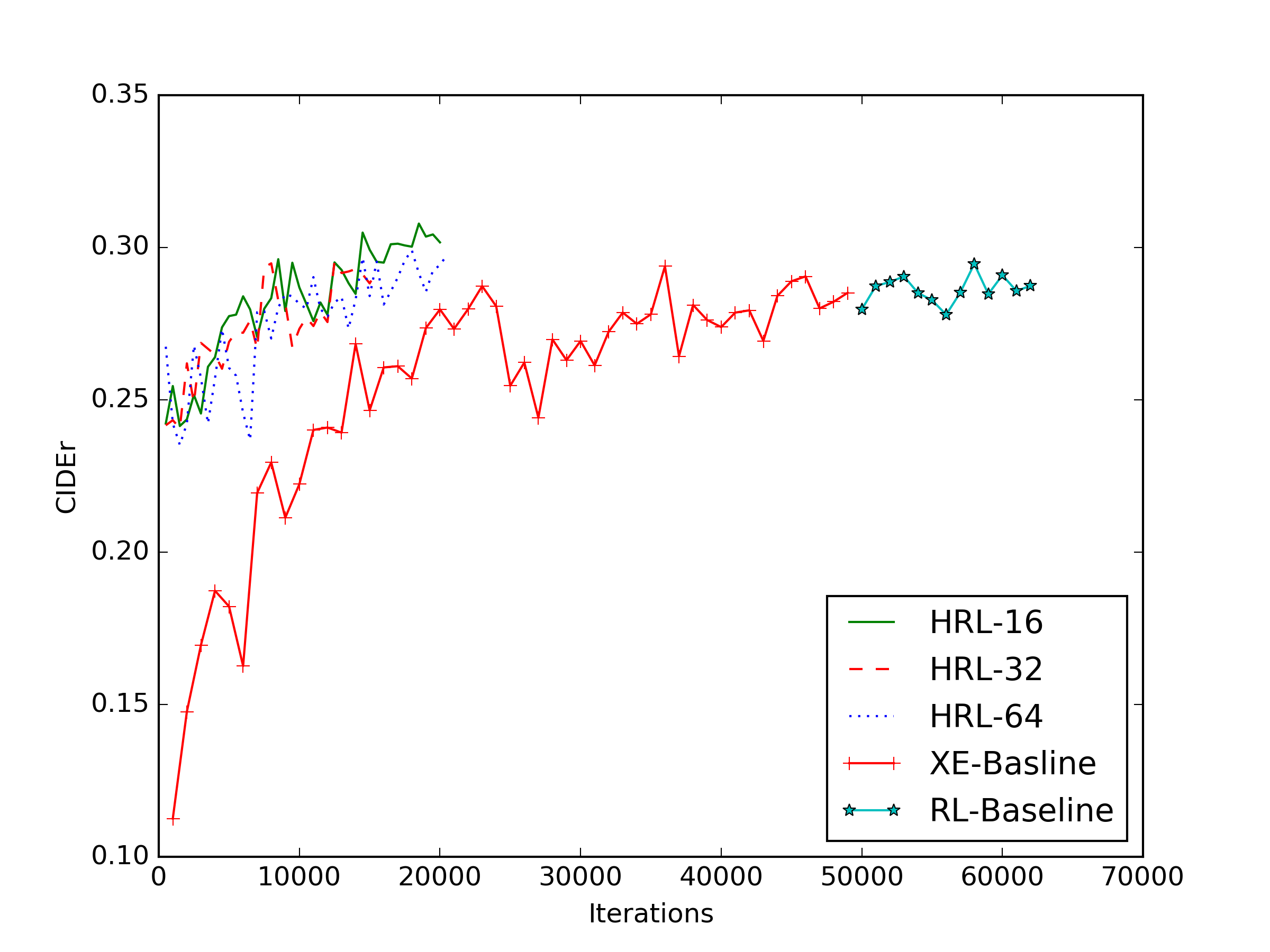}  
\end{center}
\vspace*{-1ex}
   \caption{Learning curves of the CIDEr scores of different captioning models, including XE-baseline, RL-baseline and HRL models with goal dimension of 16, 32 and 64.}
\label{fig:curve}
\end{figure}

\paragraph{Learning Curve}
For a more intuitive view of the models, we drew the learning curves of the CIDEr scores on validation set (see Figure~\ref{fig:curve}). Note that the RL-baseline model was first warmed up with cross-entropy loss, and then improved using the REINFORCE algorithm. Particularly, after we trained the XE-baseline model, we switched to policy gradient and continued training the RL-baseline model on it. HRL models were resumed training on a shorter warm-start period. As is shown in Figure~\ref{fig:curve}, the HRL models converged faster and achieved better peak points than the baseline methods. HRL-16 reached the highest point. 

\section{Conclusion}
In this paper, we propose a hierarchical reinforcement learning framework for video captioning, which aims at improving the fine-grained generation of video descriptions with rich activities. 
Our HRL model obtains the state-of-the-art performance on both the widely used MSR-VTT dataset and the newly introduced Charades Captions dataset for fine-grained video captioning.

In the future, we plan to explore the attention space and utilize features from multiple modalities to boost our HRL agent. We believe that the results of our method can be further improved by employing different types of features, \textit{i.e.} C3D features~\cite{tran2015learning}, optical flows, etc. Meanwhile, we will investigate the HRL framework in other similar sequence generation tasks like video/document summarization. 

\section*{Acknowledgement}
We would thank Ramakanth Pasunuru and Ruotian Luo for clarifying the technical details of their paper/code, and Wenhan Xiong for his help on debugging the model. Personally, Xin would appreciate the care from his girlfriend (now his wife) when he was busying working on the paper.

\newpage
\appendix

\section*{Supplementary Material}

\section{Attention Visualization}
\begin{figure}[t]
\begin{center}
\includegraphics[width=3.4in]{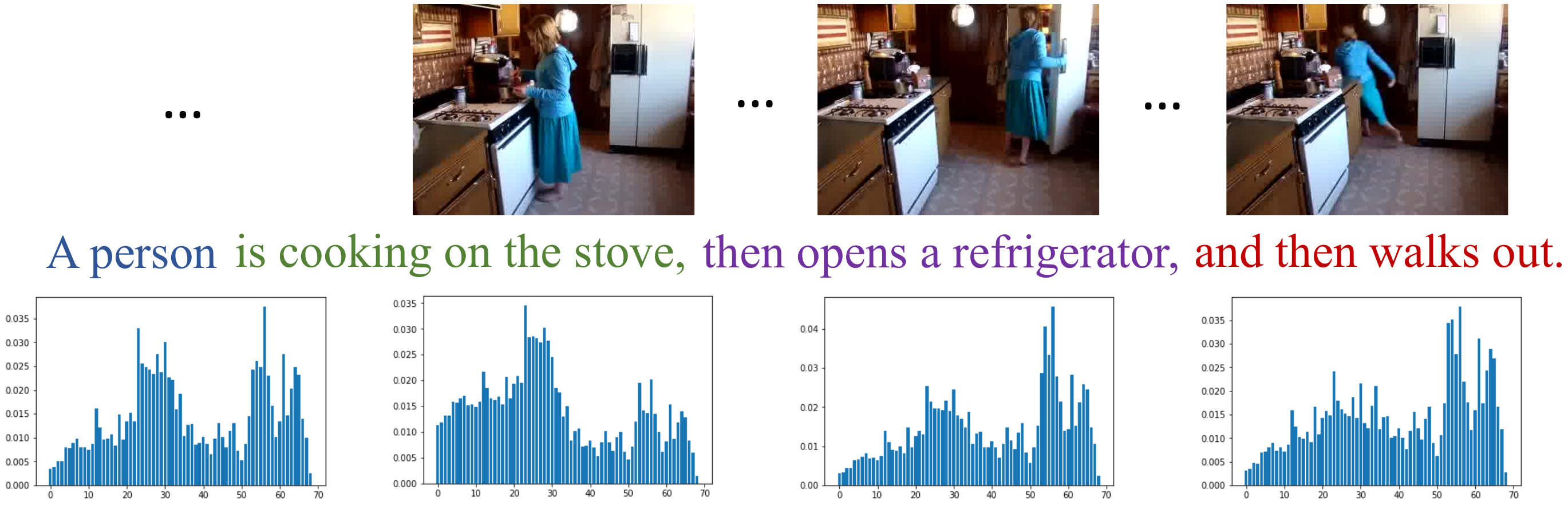}  
\end{center}
   \caption{A visualization demo of the attentions. Different text segments were in different colors, and the associated attentions were provided below the corresponding segments. We also showed the keyframe in the top row, which was selected from the most noticeable area for each segment.}
\label{fig:vis}
\end{figure}

Fig.~\ref{fig:vis} demonstrated a visualization example where the associated attentions of the learned text segments over video frames were plotted. Clearly, when generating different text segments, the HRL model attended to different temporal frames. For example, when the model was producing the segment \textit{is cooking on the stove}, the first halve of the video, which contained the action \textit{cooking}, played a more important role with larger attention values.

\section{Qualitative Examples on MSR-VTT}
\begin{figure*}[t]
\begin{center}
\includegraphics[width=6.8in]{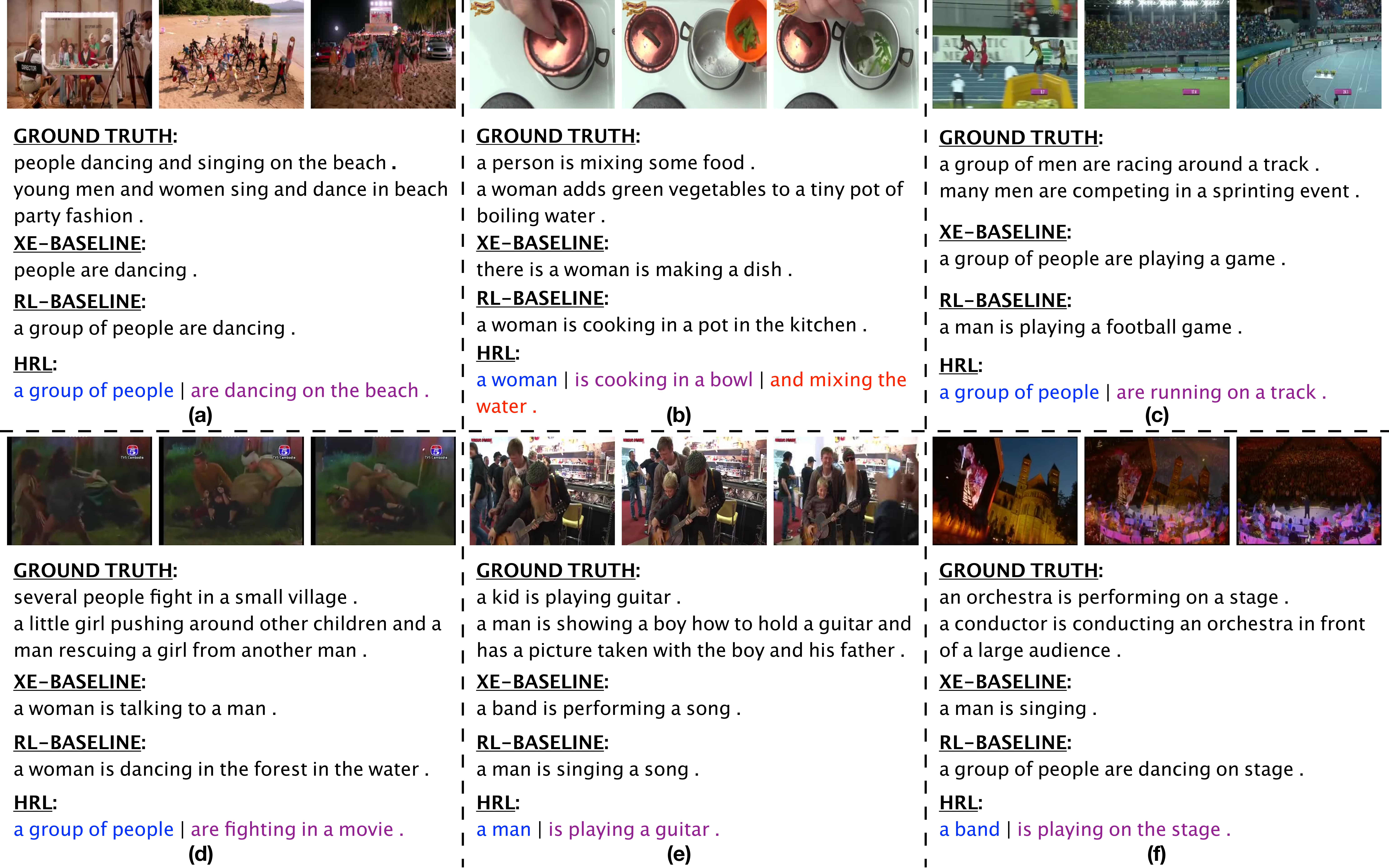}  
\end{center}
   \caption{Qualitative comparison with the baseline methods on MSR-VTT dataset. For each video example, we listed two ground truth captions, the generated result by XE-baseline (cross entropy), the result by RL-baseline (policy gradient), and the result by our HRL method (hierarchical reinforcement learning). In our HRL results, different segments were in different colors and separated with $``|"$.}
\label{fig:msrvtt_case_study}
\end{figure*}

In the main paper, we showed some generated results on Charades Captions dataset. Here we demonstrated more qualitative examples on MST-VTT dataset in Figure~\ref{fig:msrvtt_case_study}.

Particularly, Example (a) and (b) revealed that our HRL method was able to capture more details of the video content and generate more fine-grained descriptions. For example, our HRL model provided both the event (\textit{a group of people are dancing}) and the scene (\textit{on the beach}) in Example (a) while the other baseline methods failed to depict where the event is happening.
Example (c) (d) (e) and (f) further illustrated the correctness and accuracy of our HRL results. For instance, in Example (c), only the result of our HRL method described the video correctly. The ground truth caption was \textit{a group of men are racing around a track} and our result was \textit{a group of people are running on a track}. While both the XE-baseline and RL-baseline captioned by mistake the video with \textit{a group of people are playing a game} and \textit{a man is playing a football game} respectively.  
Apparently, compared with the results of the baseline methods, our results were more accurate and descriptive in general.

\section{Network Architecture}
\label{sec:network}
In this section, we illustrate the exact architecture used for the experiments (see Figure~2 in the main paper). 

\paragraph{Encoders} 
For both datasets, we sampled each video at $3 fps$ and used ResNet-152~\cite{he2016deep} (pretrained CNN model on ImageNet) to extract frame features without fine-tuning. Then the 2048-dim frame features were projected to 512-dim. The low-level encoder was a Bi-LSTM with hidden size 512, and the high-level encoder was an LSTM with hidden size 256. 

\paragraph{Worker} 
The worker network consisted of a worker LSTM with hidden size 1024, an attention module similar to the one proposed by Bahdanau \textit{et al.}~\cite{bahdanau2014neural}, a word embedding of size 512, and a projection module (Linear $\rightarrow$ Tanh $\rightarrow$ Linear $\rightarrow$ SoftMax) that produced the probabilities over all tokens in the vocabulary. 

\paragraph{Manager}
The manager network was composed of a manager LSTM with hidden size 256, an attention module, and a linear layer that projected the output of the LSTM into latent goal space.

\paragraph{Internal Critic} The internal critic was also an RNN network, which contained a GRU~\cite{cho2014learning}, a built-in word embedding, a linear layer, and a Sigmoid function. The hidden size of the GPU and the word embedding size were both 128 for MSR-VTT and 64 for Charades Captions.

\section{Training Details}
All the hyperparameters were tuned on the validation set, including the dimension sizes in Sec.~\ref{sec:network}. Moreover, we adopted Dropout~\cite{srivastava2014dropout} with a value 0.5 for regularization. All the gradients were clipped into the range [-10, 10]. We initialized all the parameters with a uniform distribution in the range [-0.1, 0.1]. For MSR-VTT dataset, we used a fixed step size of 50 for the encoder LSTMs and a maximum length of 30 for the captions. For Charades Captions dataset, they were set to 150 and 60 respectively.

To train the cross-entropy (XE) models, Adadelta optimizer~\cite{zeiler2012adadelta} was used with batch size 64. The learning rate was initially set as 1 and then reduced by a factor 0.5 when the current CIDEr score did not surpass the previous best for 4 epochs. Schedule sampling~\cite{bengio2015scheduled} was employed to train the XE models.
When training the RL and HRL models, we used the pretrained XE models to warm start and then continued training them with a learning rate 0.1. 
The discounted factors of the Manager and the Worker were both 0.95. At test time, we used beam search of size 5. 

{\small
\bibliographystyle{ieee}
\bibliography{egbib}
}

\end{document}